\pdfoutput=1
\documentclass[11pt]{article}

\usepackage[preprint]{acl}

\usepackage{times}
\usepackage{latexsym}

\usepackage[T1]{fontenc}
\usepackage[utf8]{inputenc}

\usepackage{microtype}

\usepackage{inconsolata}

\usepackage{graphicx}

\usepackage{multicol}
\usepackage{multirow}
\usepackage{tcolorbox}
\usepackage{booktabs}
\usepackage{amsmath}
\usepackage{amssymb}
\usepackage{amsthm}
\usepackage{mathtools}
\usepackage{cleveref}
\usepackage{tikz}
\usetikzlibrary{positioning}
\usepackage{colortbl}
\usepackage{algorithm}
\usepackage[noend]{algpseudocode}

\DeclarePairedDelimiterX{\infdivx}[2]{(}{)}{#1\;\delimsize\|\;#2}

\theoremstyle{plain}

\usepackage{bm}

\DeclareMathOperator*{\argmax}{arg\,max}
\DeclareMathOperator*{\argmin}{arg\,min}

\newcommand{\st}{\text{\, s.t. \,}}

\newcommand{\mbR}{\mathbb{R}}

\newcommand{\Method}{DBD}

\title{Do More Details Always Introduce More Hallucinations in LVLM-based Image Captioning?}

\author{Mingqian Feng,  Yunlong Tang,  Zeliang Zhang,  Chenliang Xu\\
University of Rochester\\
{\tt\small \{mingqian.feng,yunlong.tang,zeliang.zhang,chenliang.xu\}@rochester.edu}
}

\begin{document}
\maketitle

\begin{abstract}
Large Vision-Language Models (LVLMs) excel in integrating visual and linguistic contexts to produce detailed content, facilitating applications such as image captioning. However, using LVLMs to generate descriptions often faces the challenge of object hallucination (OH), where the output text misrepresents actual objects in the input image. While previous studies attribute the occurrence of OH to the inclusion of more details, our study finds technical flaws in existing metrics, leading to unreliable evaluations of models and conclusions about OH. This has sparked a debate on the question: \textit{Do more details always introduce more hallucinations in LVLM-based image captioning?}

In this paper, we address this debate by proposing a novel decoding strategy, Differentiated Beam Decoding (DBD), along with a reliable new set of evaluation metrics: CLIP-Precision, CLIP-Recall, and CLIP-F1. DBD decodes the wealth of information hidden in visual input into distinct language representations called unit facts in parallel. This decoding is achieved via a well-designed differential score that guides the parallel search and candidate screening. The selected unit facts are then aggregated to generate the final caption. Our proposed metrics evaluate the comprehensiveness and accuracy of image captions by comparing the embedding groups of ground-truth image regions and generated text partitions. Extensive experiments on the Visual Genome dataset validate the effectiveness of our approach, demonstrating that it produces detailed descriptions while maintaining low hallucination levels.
\end{abstract}

\section{Introduction}
Large Vision-Language Models (LVLMs)~\cite{liu2023improved,long2022vision, zhu2023minigpt,openai2024gpt4} have been broadly employed in a range of multimodal applications, such as image captioning~\cite{li2022blip, liu2023visual, chen2023shikra, wang2023caption, xuan2023pink}, where the objective is to generate textual descriptions that accurately encapsulate visual data. On the one hand, LVLMs are expected to generate descriptions that are detailed and comprehensive rather than overly succinct, ensuring pivotal visual information is not omitted. On the other hand, detailed and long descriptions from these LVLMs often suffer from a significant challenge known as \textit{hallucination}~\cite{rohrbach2018object}, where output text is semantically coherent but misaligned with the actual objects present in the input image. In this work, we mainly consider \textit{object hallucination}. Such issues, whether of omission or hallucination, limit practical applications in safety-critical fields such as medical imaging~\cite{hu2023advancing,alsharid2022gaze,alsharid2021acourse} or autonomous driving~\cite{chen2023driving,jin2023adapt}, where accurate and comprehensive outputs are essential.

\begin{figure}[t]
    \centering
    \includegraphics[width=\linewidth]{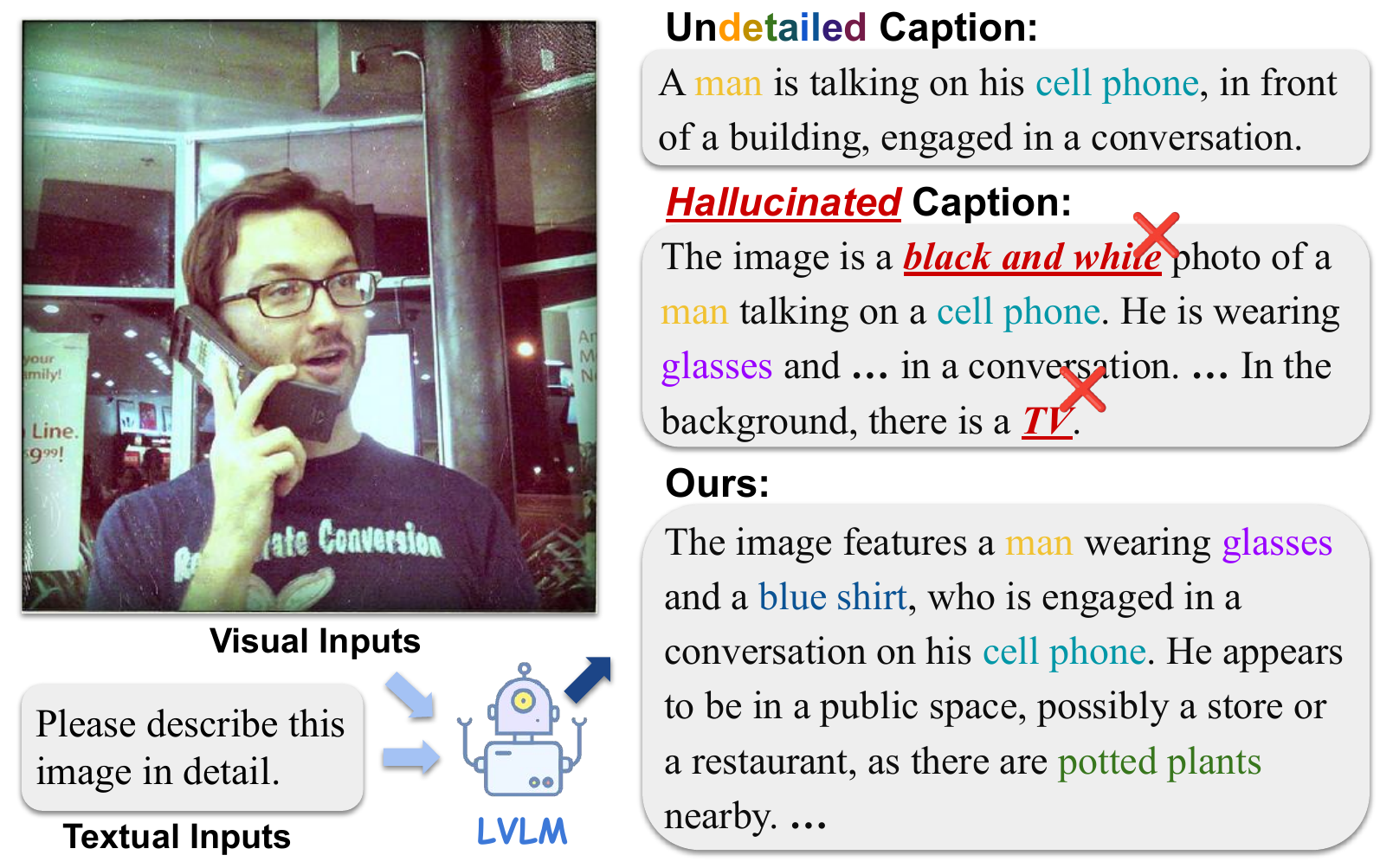}
    \caption{Comparison of captions. The top caption ignores much visual information. The middle one includes details but also introduces hallucinations. Our method (bottom) provides detailed and accurate captions.}
    \label{fig:teaser_fig}
    \vspace{-6mm}
\end{figure}

As shown in \Cref{fig:teaser_fig}, a short caption accurately describes the main objects, \textit{man} and \textit{cell phone}, in the center of the image but overlooks details such as the man's \textit{glasses} and the \textit{potted plant} in the corner. However, when we encourage LVLMs to generate a longer caption to include more image details, we can see the occurrence of hallucinations, such as a nonexistent object, \textit{TV}, and inaccurate attributes, \textit{black and white}. A question naturally arises: \textit{\textbf{Do more details always introduce more hallucinations in LVLM-based Image Captioning?}}

There remains a debate about answering this question. On the one hand, substantial evidence ~\cite{zhou2023analyzing,deng2024seeing} indicates that hallucinations are more likely to appear in later sentences of long text sequences where models tend to describe details. Moreover, the native design of the model results in the subsequent description of the details receiving more interference than the previous text, which is misleading from the previous text itself. On the other hand, our in-depth manual analysis reveals that existing metrics overstate the presence of hallucinations in detailed descriptions. This inaccuracy stems from two main factors: first, human-generated ground-truth captions often lack detail; second, rule-based evaluations fail to grasp the true intent of image captioning, sometimes even contradicting it. The more detailed the description, the more serious the impact of these two factors, rendering the evidence less credible. 
% We discuss the arguments on both sides at length in \Cref{sec:det_hall_debate}.

In this work, we aim to \textit{\textbf{settle down this debate by demonstrating that, with proper guidance, LVLMs can produce detailed descriptions while maintaining low hallucination levels.}} To achieve this goal, we propose a novel decoding strategy, Differentiated Beam Decoding (DBD), along with a new set of evaluation metrics: CLIP-Precision, CLIP-Recall, and CLIP-F1. Our proposed DBD is grounded in parallel decoding to eliminate misleading information from pre-generated texts and differentiated searching to compel the LVLM to describe the image from various perspectives. 
It operates in three phases: differentiated parallel search, post-search selection, and unit facts summarization.
Our introduced metrics utilize different divisions of the image and the caption to obtain a group of embeddings, respectively. The percentage that one group of embeddings can express the other is calculated, serving as CLIP-precision and CLIP-recall, which are then combined to produce CLIP-F1.

Our contributions are fourfold:

\noindent $\bullet$ We introduce a novel, training-free decoding method that generates in parallel multiple facts of distinct aspects of an image and synthesizes them into a comprehensive and accurate description.

\noindent $\bullet$ We conduct an in-depth analysis of widely used hallucination metrics, revealing that over $50\%$ of identified object hallucinations are unjust.

\noindent $\bullet$ We propose a novel set of metrics that specifically and separately evaluate the comprehensiveness and hallucination level of image captioning.

\noindent $\bullet$ 
We perform extensive experiments to validate the effectiveness of our approach. We tested various LVLMs on the Visual Genome dataset using the proposed metric. 
The results demonstrate that our method produces more detailed image descriptions while maintaining a low level of hallucinations.

\section{Related Work}

\subsection{LVLM-based Image Captioning }

Image captioning~\citep{bai2018survey} aims to bridge the gap between visual perception and language understanding. Compared with traditional models~\citep{vinyals2015show,wang2020overview}, by training on extremely large-scale datasets, Large Vision-Language Models (LVLMs) can give us more detailed and contextually rich descriptions to improve the performance of image understanding.  Representative works include LLaVA-1.5~\cite{liu2023visual}, MiniGPT-4~\cite{zhu2023minigpt}, and mPLUG-Owl2~\cite{ye2023mplug}. 

However, the use of LVLMs in the image captioning task sometimes comes with hallucination~\citep{gunjal2024detecting}, which is an issue in that LVLMs generate descriptions containing objects or details not present in the image. Existing methods to mitigate hallucinations include reinforced grounding~\cite{favero2024multi}, adversarial training~\cite{park2024mitigating}, and visual verification steps~\cite{liu2023aligning, lu2023evaluation}.

% Hallucination is a significant issue where LVLMs generate descriptions containing objects or details not present in the visual input. This problem is exacerbated by reliance on language patterns and biases from large pretraining datasets. Existing methods to mitigate hallucinations include reinforced grounding techniques, adversarial training, and visual verification steps~\cite{liu2023aligning, lu2023evaluation}. To improve caption details, approaches that leverage granular visual features and context-aware language models have been implemented. Hierarchical attention mechanisms help generate specific and contextually rich descriptions by focusing on finer details~\cite{fu2016aligning}. Incorporating external knowledge bases and contextual embeddings can further enrich captions~\cite{yin2019context}. Multi-stage training that refines coarse descriptions also improves detail~\cite{gu2018stack}. 
% In this work, we propose an approach that explores distinct narrative directions in parallel, enhancing the richness and reducing hallucinations. 

\subsection{Evaluation Metrics of Image Captioning}
Three metrics are widely used to evaluate the performance of  the image captioning task, including Caption Hallucination Assessment with Image Relevance (CHAIR)~\cite{rohrbach2018object}, Bilingual Evaluation Understudy (BLEU)~\cite{papineni2002bleu}, and the CLIP Score~\cite{hessel2022clipscore}. 

CHAIR evaluates object hallucination in image captioning, consisting of (1) CHAIR\(_S\), measuring the proportion of sentences with hallucinated objects, and (2) CHAIR\(_I\), assessing the proportion of hallucinated objects among all mentioned objects.  BLEU evaluates the match between generated and reference texts based on n-gram precision. 
The reference-free CLIP Score is also used to evaluate image captioning.
However, each of these metrics has limitations. CLIP Score lacks detail, which is crucial for high-quality image captioning. CHAIR and BLEU metrics are often limited by humans' undetailed ground-truth caption annotations. 
% These challenges highlight the need for more robust and detailed evaluation metrics in captioning.

\section{Detail-Hallucination Debate} \label{sec:det_hall_debate}

\subsection{Co-occurrence of Detail and Hallucination}
The arguments regarding the co-occurrence of detail and hallucination fall into two main points.

\begin{figure}[t]
    \centering
    \includegraphics[width=0.99\linewidth]{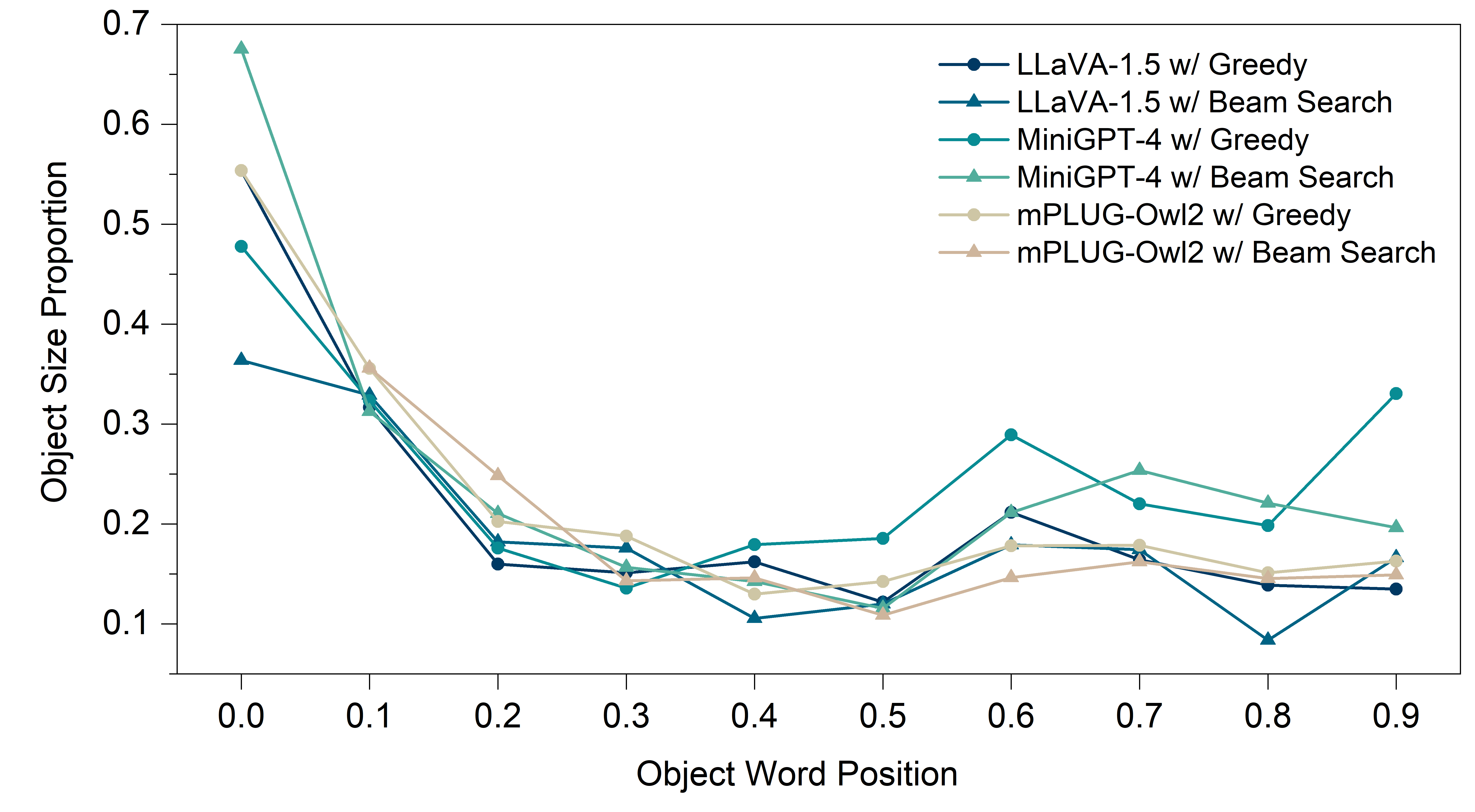}
    \caption{Object word position in the caption v.s. object size proportion to the image. As the caption progresses, LVLMs are prone to describe smaller objects.}
    \label{fig:det_app_later}
    \vspace{-4mm}
\end{figure}

First, substantial experimental evidence from prior research indicates that hallucinations tend to appear more frequently in the later parts of captions. For instance, Figure 2a. in \citet{deng2024seeing} shows a consistent pattern across multiple LVLMs, where the probability of hallucination increases as the sentence progresses. Additionally, Figure 1c. in \citet{zhou2023analyzing} highlights that the end of descriptions is a predominant high-density area for hallucinatory objects. Although this observation is intuitive, our further analysis of the LVLM-based captioning demonstrates that models tend to describe details in the later part of the caption, as presented in \Cref{fig:det_app_later}. 
We provide the settings for this experiment and more discussions in the \Cref{apdx:det_hall_debate}.
% Together, this evidence suggests the co-occurrence of details and hallucinations in LVLM-based image captioning.

\begin{figure}[t]
    \centering
    \includegraphics[width=\linewidth]{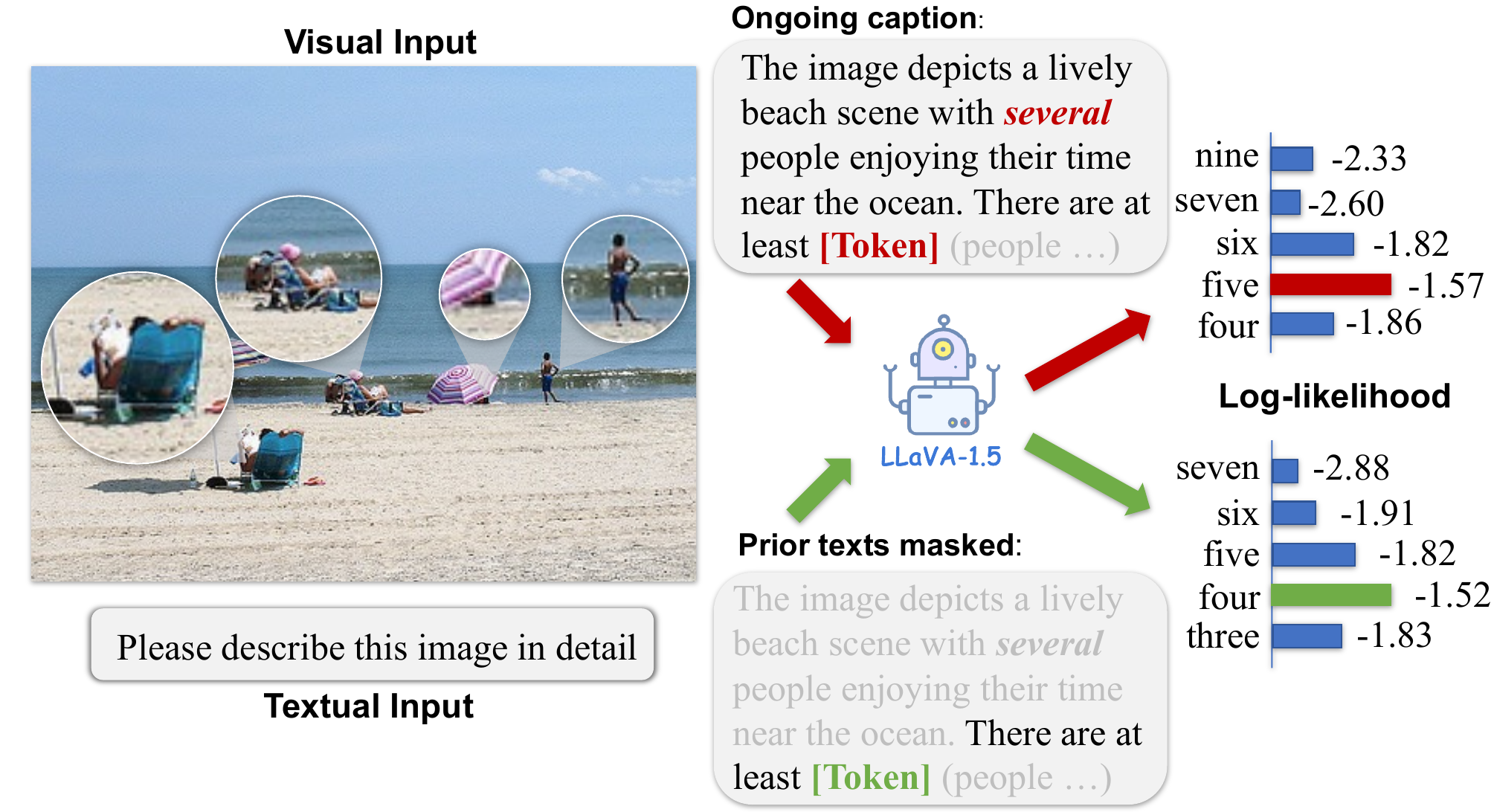}
    \caption{Misleading from pre-generated texts. With prior output engaged, LLaVA-1.5 predicts most likely \textit{five} people in the scene. However, with the first sentence masked during the coding, the number decreases to \textit{four}.}
    \label{fig:motivation_misleading}
    \vspace{-4mm}
\end{figure}

Second, the inherent attention mechanism of looking back causes the subsequent detailed description to receive more interference. In a typical sequential generation, it is reasonably expected that earlier text influences later text, and a rational consensus is that this influence should only affect the direction of the narrative, not the critical factual content. However, in vision-language tasks such as image captioning, where complex information is hidden in the visual embedding rather than explicitly expressed in language, the previous descriptions can interfere with the generation of subsequent details. For example, as illustrated in \Cref{fig:motivation_misleading}, in the original caption, LLaVA-1.5 initially identifies \textit{five} people in the scene. However, when we mask the pre-generated texts during inference, the number decreases to \textit{four}, indicating the misleading from the prior words such as \textit{``several.''}

\subsection{Inflated Hallucination Presence}
Nevertheless, our in-depth analysis of the classic hallucination evaluation method, CHAIR, reveals that the presence rate of hallucination is inflated, particularly in detailed captions, refuting the experimental evidence from the co-occurrence side. 

Specifically, we generated captions for $500$ randomly selected images from the MSCOCO validation dataset using LLaVA-1.5~\cite{liu2023visual} and evaluated them using the CHAIR~\citep{rohrbach2018object} framework, a widely used method in the literature. Upon manually inspecting the samples identified as hallucinated, we found that over $55\%$ of these identified hallucinations are unjustified. The misjudgment is due to two main reasons, and both are exacerbated in detailed captions.

\begin{figure}[t]
    \centering
    \includegraphics[width=\linewidth]{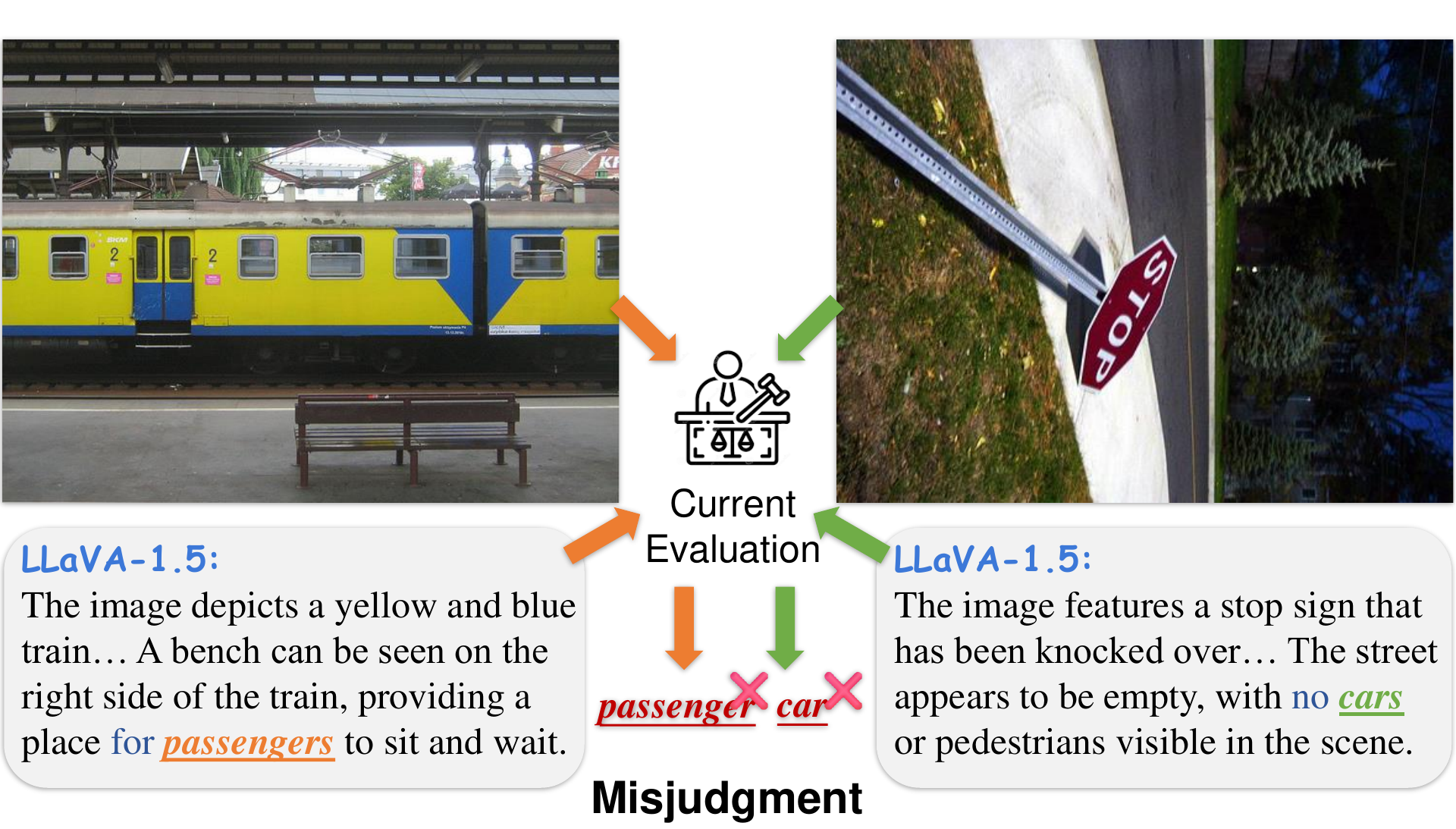}
    \caption{Misjudgement from rule-based identification. The current framework misunderstands elaborate captions, where many object words appear not for existence but for decorative features or even for non-existence.}
    \label{fig:misjudg_misund}
    \vspace{-6mm}
\end{figure}

First, the rule-based object extraction is rigid and prone to errors, leading to a misunderstanding of the caption. The current rule assumes that if an object word appears in the caption, the description expresses the presence of the object in the image. However, detailed captions often include decorative text that contains object words without indicating their presence. For instance, as shown in \Cref{fig:misjudg_misund}~(left), the object word, \textit{passenger}, is used to describe a characteristic of the seat, for passengers to sit and wait, not to indicate the presence of a passenger.  Similarly, detailed captions might explicitly state the absence of certain \textit{objects} related to the scene, such as \textit{car} in \Cref{fig:misjudg_misund}~(right), yet are still labeled as hallucinations. Other inaccurate extractions include identifying \textit{orange} from \textit{orange juice}, \textit{bowl} from \textit{toilet bowl}, and so on.

\begin{figure}[t]
    \centering
    \includegraphics[width=\linewidth]{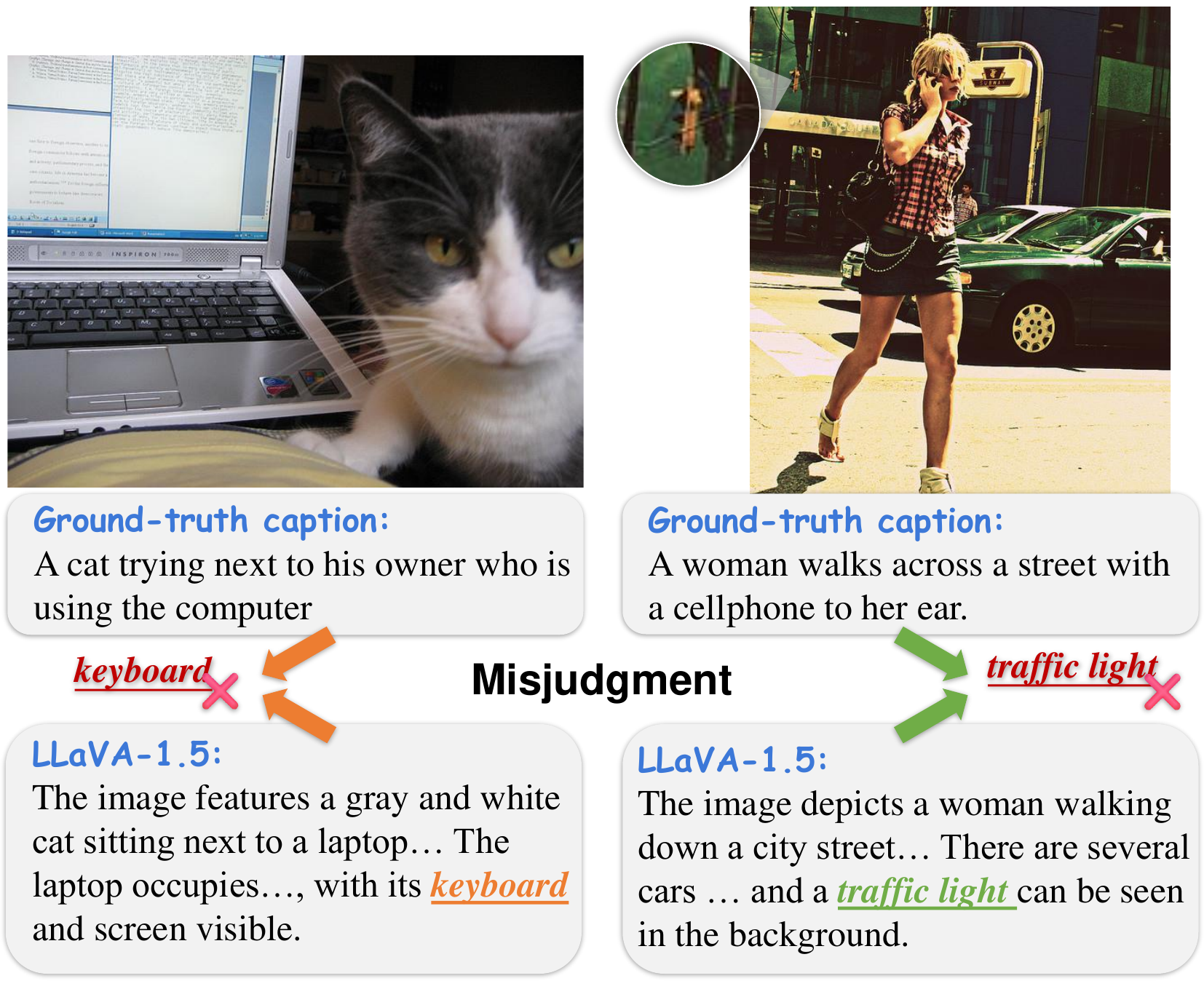}
    \caption{Misjudgement from undetailed ground truths. The current framework identifies hallucination based on language ground truths, restricting the capability to evaluate captions in more detail than ground truths. }
    \label{fig:misjudg_undetailed_gt}
    \vspace{-4mm}
\end{figure}

Second, the language-reference-required comparisons are restricted by the quality of annotations. 
Any ground-truth caption is essentially a lossy compression of the original visual information, which falls short when more detailed predictions arise. This phenomenon is inevitable, akin to our perpetual quest for additional decimal places in $\pi$.
% While it is reasonable to refer to ground-truth captions, this is based on the assumption that the model's predictions are of lower quality than theirs. However, with the growth of LVLMs' capability, the existing caption annotation has failed to meet this condition, especially in the widely used MSCOCO dataset. 
This issue is particularly prevalent in the current evaluation framework that widely uses MSCOCO. For example, as presented in \Cref{fig:misjudg_undetailed_gt}, ground-truth captions from humans often omit subassemblies, like \textit{keyboard} of a laptop or small details such as \textit{traffic light}. We remark that the ground truths
shown in \Cref{fig:misjudg_undetailed_gt} are obtained by summarizing all human captions provided by MSCOCO. Additionally, human annotations frequently ignore the background environment, such as \textit{table}, where the objects are situated.
Consequently, many accurate details are incorrectly recognized as hallucinations. We provide more examples of failure in \Cref{apdx:det_hall_debate}.

% \subsection{Statement of Purpose}
% Our goal is to settle this detail-hallucination debate and suggest that with proper guidance, LVLMs can produce detailed descriptions while maintaining low hallucination levels. We first propose a novel decoding strategy that eliminates the extra interference from pre-generated text and provides detailed distinct unit facts of the image. We then introduce a new set of metrics that addresses the problems mentioned above and conduct extensive experiments to evaluate our method.

\begin{tcolorbox}[colframe=brown,colback=white, title = {Our arguments}]
\noindent\textbf{Mitigating interference}. While the validity of experimental evidence that highlights the co-occurrence of detail and hallucination using current evaluation metrics has been challenged, the inherent problem posed by the attention mechanism's design persists. To mitigate the issue of the extra interference from pre-generated text and allow for high-fidelity and detailed image captioning, we introduce a novel decoding method, Differentiated Beam Decoding (DBD).

\noindent\textbf{Enhancing evaluation metrics}. Despite a new method to reduce interference issues, the current metrics still have significant limitations, as emphasized by the opposing viewpoint. To provide a more comprehensive and reliable evaluation of both the thoroughness of detail and the accuracy of content, we propose a novel set of metrics, CLIP-Recall, -Precision, and -F1.
\end{tcolorbox}

\section{Methodology}
% To settle this detail-hallucination debate, we first introduce our method, Differentiated Beam Decoding (DBD). DBD mitigates the issue of the extra interference from pre-generated text, allowing for high-fidelity and detailed image captioning. 

\subsection{Preliminary} \label{sec:method_preliminary}
\noindent\textbf{LVLMs}. We consider a Large Vision-Language Model (LVLM), denoted as $\mathcal{M}$, which receives an image $v$ and a text prompt $x$, including the pre-generated tokens, to infer the next token probability. This process is formalized as follows:
\begin{equation*}
    \setlength\abovedisplayskip{3pt}
    \setlength\belowdisplayskip{3pt}
    \mathcal{M}(\bm{v},\bm{x},\bm{y}_{<t})\coloneqq p_{\mathcal{M}}(\cdot \mid \bm{v},\bm{x},\bm{y}_{<t}).
\end{equation*}
Given each token's probability from the LVLM, the sentence-level log-likelihood is defined as:
\begin{equation*}
    \setlength\abovedisplayskip{3pt}
    \setlength\belowdisplayskip{3pt}
    \log p_{\mathcal{M}}(\bm{y}_{\le t} \mid \bm{v},\bm{x}) = \sum_{i} \log p_{\mathcal{M}}(\bm{y}_{i} \mid \bm{v},\bm{x},\bm{y}_{<i}).
\end{equation*}

\noindent\textbf{Decoding strategy}. Decoding strategies dictate how to harness the next token probability to select the next token $y_t$ from the set $S$ of all possible tokens in the vocabulary, which includes special tokens such as <BOS>, <EOS>, and <PAD>. Greedy decoding, the simplest strategy, selects the next token with the highest probability, formalized as 
\begin{equation*}
    \setlength\abovedisplayskip{3pt}
    \setlength\belowdisplayskip{3pt}
    y_t = \argmax_{s\in S} \log p_{\mathcal{M}}(s \mid \bm{v},\bm{x},\bm{y}_{<t}).
\end{equation*}
% The decoding terminates when either a <EOS> token is selected or the maximum length $\Bar{L}$ is reached. 

\begin{figure*}[t]
    \centering
    \includegraphics[width=\linewidth]{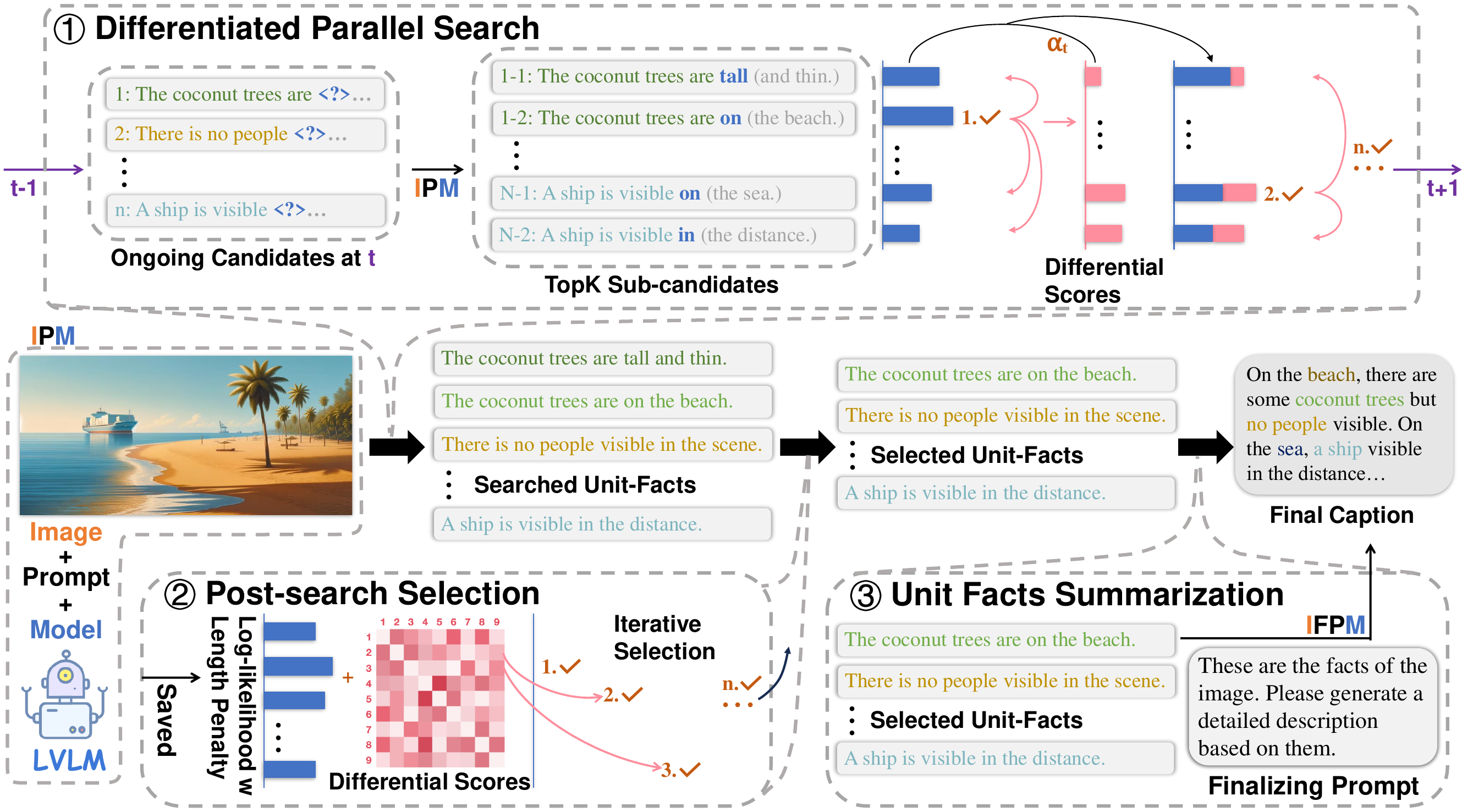}
    \caption{Overview of \Method{} method. In the Differentiated Parallel Search phase, we balance log-likelihoods from the model and differential scores to search distinct unit facts. In the Post-search Selection phase, we select multiple representative unit facts using length-penalized log-likelihood and differential scores. In the Unit Facts Summarization phase, we task the LVLM to generate the final caption based on selected unit facts and the image.}
    \label{fig: pipeline}
    \vspace{-4mm}
\end{figure*}

\subsection{Differentiated Beam Decoding} \label{sec:method_DBD}
As depicted in \Cref{fig: pipeline}, our method, Differentiated Beam Decoding (\Method{}), comprises three phases. First, we search for multiple candidates of the image's unit facts in parallel, ensuring there is no precedence and no extra interference. Subsequently, we select a subset from these unit-fact candidates to eliminate redundancy and improve efficiency. Finally, we task the LVLM to finalize the caption based on selected unit facts and the image.

\noindent\textbf{Differential score}.
Before detailing the three phases, we introduce the sequence-level and set-level \textit{Differential Score}, which are utilized in the search and selection phases to promote diversity.
% To promote diversity in search and selection, we introduce the \textit{Differential Score}.
% We define the sequence-level and set-level \textit{Differential Score} as follows.  

For two sequences, $\bm{y}^{(1)}=(y^{(1)}_1,...y^{(1)}_{L_1})$ and $\bm{y}^{(2)}=(y^{(2)}_1,...y^{(2)}_{L_2})$, the sequence-level differential score is computed as the negative cosine similarity between their tokens' average hidden state vectors from the LVLM, or equivalently,
\begin{equation*}
    \setlength\abovedisplayskip{3pt}
    \setlength\belowdisplayskip{3pt}
    d_{\mathcal{M}}(\bm{y}^{(1)},\bm{y}^{(2)}) = -\frac{1}{L_1 L_2} \sum_{i=1}^{L_1}\sum_{j=1}^{L_2} \cos(h^{(1)}_i,h^{(2)}_j),
\end{equation*}
where $h_i$ denotes the hidden states of the token $y_i$. 

For a set of candidates $\mathcal{Y} = \{\bm{y}^{(1)},...\bm{y}^{(k)}\}$, the set-level differential score, $D_{\mathcal{M}}(\mathcal{Y})$ aggregates the sequence-level differential scores across all unique candidate pairs $(\bm{y}^{(i)},\bm{y}^{(j)})$ for $i\neq j$, as follows:
\begin{equation*}
    \setlength\abovedisplayskip{3pt}
    \setlength\belowdisplayskip{3pt}
    D_{\mathcal{M}}(\mathcal{Y}) = \sum_{i=2}^k \sum_{j=1}^{i-1} d_{\mathcal{M}}(\bm{y}^{(i)},\bm{y}^{(j)}).
\end{equation*}

\noindent\textbf{Differentiated parallel search}. In this phase, we leverage differential scores to guide the generation of independent and distinct unit facts in parallel. 

Generating fluent, accurate, and varied descriptions of images requires balancing high model probability with significant differential scores. Our objective is to identify the optimal set of unit facts, $\mathcal{Y}=\{\bm{y}^{(1)},...\bm{y}^{(n)}\}$, that contains $n$ unit facts and maximizes a combination of the log-likelihoods and a set-level weighted differential score,
\begin{equation*}
    \setlength\abovedisplayskip{3pt}
    \setlength\belowdisplayskip{3pt}
    \max_{\mathcal{Y}} [\sum_{i=1}^n \log p_{\mathcal{M}}(\bm{y}^{(i)}\mid \bm{v},\bm{x}) + \alpha D_{\mathcal{M}}(\mathcal{Y})]
\end{equation*}
where $n$ is the predefined size of the candidate set and $\alpha$ is a weight that balances the impact of the differential score. Directly addressing this optimization is complex, so we adopt an iterative approach to resolve the sub-problems for $i\in[n]$,
\begin{equation*}
    \setlength\abovedisplayskip{3pt}
    \setlength\belowdisplayskip{3pt}
    \max_{\mathcal{Y}} [\log p_{\mathcal{M}}(\bm{y}^{(i)}\mid \bm{v},\bm{x}) + \sum_{j=1}^{i-1} d_{\mathcal{M}}(\bm{y}^{(i)},\bm{y}^{(j)})].
\end{equation*}

The concrete decoding procedure is similar to the Beam Search but incorporates dynamic criteria based on the differential scores of pre-picked candidates. We provide more discussions on beam search in \Cref{apdx:disc_beam_subsection}. Specifically, at each step $t$, we maintain a set of parallel candidates $\mathcal{Y}_t=\{\bm{y}^{(1)}_{\le t},...,\bm{y}^{(n)}_{\le t}\}$, with each candidate representing a unit fact in progress. At the next step $t+1$, for each candidate $\bm{y}^{(i)}_{\le t}$, we obtain the top $K$ most probable next tokens predicted by the model, where $K$ is a hyperparameter. These total $nK$ next tokens, along with their source candidates, form a sub-candidate set $\Tilde{\mathcal{Y}}_{t+1}=\{\Tilde{\bm{y}}^{(1)}_{\le t+1},...,\Tilde{\bm{y}}^{(nK)}_{\le t+1}\}$. 

From this sub-candidate set, we pick next-step candidates, $\mathcal{Y}_{t+1}$. Initially, the sub-candidate with the highest sentence-level log-likelihood is chosen first and removed from the sub-candidate set:
\begin{equation*}
    \setlength\abovedisplayskip{3pt}
    \setlength\belowdisplayskip{3pt}
    \bm{y}^{(1)}_{\le t+1} = \argmax_{\Tilde{\bm{y}}_{\le t+1} \in \Tilde{\mathcal{Y}}_{t+1}} \log p_{\mathcal{M}}(\Tilde{\bm{y}}_{\le t+1} \mid \bm{v}, \bm{x}).
\end{equation*}
Subsequently, we iteratively pick and remove the sub-candidate with the highest hybrid score, defined as the log-likelihood adjusted by a weighted sum of differential scores with existing next-step candidates, from the remaining sub-candidate set. It can be formalized as follows for $2\le j \le n$:
\begin{align*}
    \bm{y}^{(j)}_{\le t+1} = &\argmax_{\Tilde{\bm{y}}_{\le t+1} \in \Tilde{\mathcal{Y}}'_{t+1}}  [  \log p_{\mathcal{M}}(\Tilde{\bm{y}}_{\le t+1} \mid \bm{v}, \bm{x}) \\
    &+  \alpha_{t+1} \sum_{k=1}^{j-1} d_{\mathcal{M}}(\Tilde{\bm{y}}_{\le t+1},\bm{y}^{(k)}_{\le t+1}) ],
\end{align*}
where $\Tilde{\mathcal{Y}}'_{t+1}$ denotes the remaining sub-candidate set, and $\alpha_{t+1}$ is a parameter that adjusts the influence of differential scores, decreasing as $t$ progresses to encourage differentiation in early stage. Suppose a picked token is the end-of-sequence token (<EOS>). In that case, this completed fact is moved to a set of final unit-fact candidates, $\mathcal{Y}_{\text{fact}}$, for the next phase, and the next sub-candidate with the highest hybrid score is picked as a replacement. Once $n$ unfinished sub-candidates are chosen, we obtain the next-step candidate set $\mathcal{Y}_{t+1}$. 

The procedure repeats until the predefined maximum time step is reached. 
% Then, all current unfinished candidates are added to the final unit-fact candidate set $\mathcal{Y}_{\text{fact}}$. Therefore, there would be at least $n$ final fact candidates in the set $\mathcal{Y}_{\text{fact}}$, and usually more. 
Throughout the process, multiple completed unit facts are added to the final unit-fact candidate set $\mathcal{Y}_{\text{fact}}$.
Typically, $\mathcal{Y}_{\text{fact}}$ contains redundancies that need to be further refined.

\noindent\textbf{Post-search selection}. In this phase, we select a subset $\Bar{\mathcal{Y}} = \{\Bar{\bm{y}}^{(1)},...\Bar{\bm{y}}^{(\Bar{n})}\}$, from the obtained facts set $\mathcal{Y}_{\text{fact}}$, to remove redundancy and improve efficiency. The goal is to identify the most representative unit facts that balance sentence-level log-likelihood with a length penalty and differential score. The selection process operates in a manner similar to the first phase. Specifically, it begins by identifying the candidate with the highest pure likelihood score. Then, it is followed by an iterative selection from the remaining facts, prioritizing those with high log-likelihoods and minimal redundancy with the already selected facts. The dynamic criterion is formalized as follows: for $j \in [\Bar{n}]$
\begin{align*}
    \Bar{\bm{y}}^{(j)} = \argmax_{\bm{y} \in \mathcal{Y}_{\text{fact}}} (& \frac{1}{|\bm{y}|} \log p_{\mathcal{M}}(\bm{y} \mid \bm{v}, \bm{x}) \\
    &+ \Bar{\alpha} \sum_{k=1}^{j-1} d_{\mathcal{M}}(\bm{y},\Bar{\bm{y}}^{(k)}) ),
\end{align*}
 where $|\bm{y}|$ represents the length of the candidate and $1/|\bm{y}|$ serves as the length penalty factor, ensuring that shorter candidates are not undervalued.

\noindent\textbf{Unit facts summarization}. In this phase, the set $\Bar{\mathcal{Y}}$ of selected unit facts are presented to the LVLM alongside the image. The model synthesizes these unit facts into a coherent final description. This step not only leverages the explicit information in the unit facts but allows for optional re-evaluation based on the image itself, ensuring that the final caption remains closely aligned with the vision.

\begin{figure}[t]
    \centering
    \includegraphics[width=\linewidth]{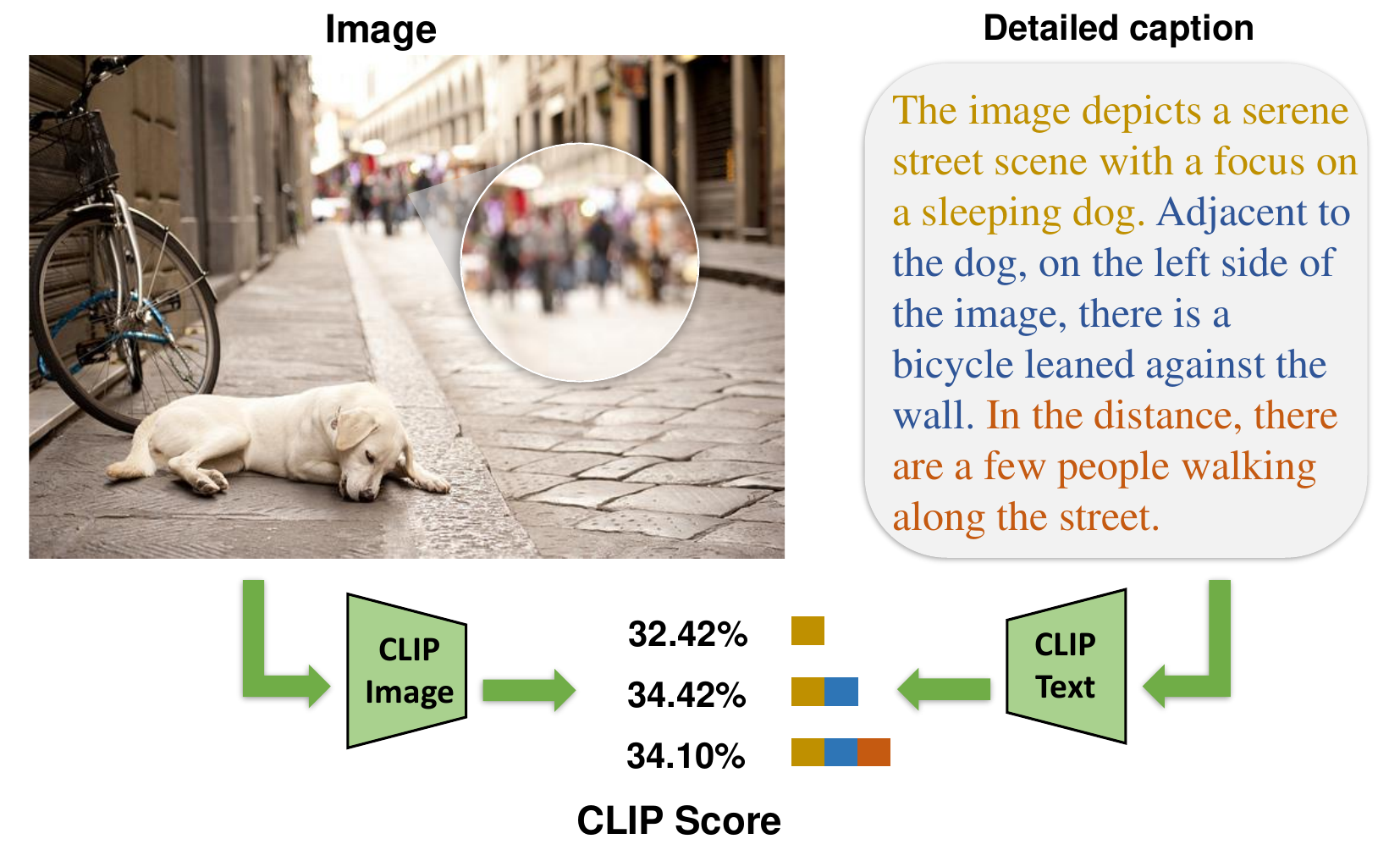}
    \caption{CLIP Scores between the image and captions with different detail levels. Each color square corresponds to a sentence of the same color. The full detailed caption has a lower CLIP score than the second one.}
    \label{fig:clip_is_biased}
    \vspace{-4mm}
\end{figure}

\subsection{CLIP-Recall, -Precision, and -F1}
Current evaluation metrics have notable limitations. BLEU~\cite{papineni2002bleu} has been unable to adapt to the diverse expression of captions from current LVLMs. The current rule-based extraction framework of CHAIR~\cite{rohrbach2018object} fails to accurately evaluate detailed captions, where many object words appear not for existence but for decorative features or even non-existence. Addressing this problem requires a deep understanding of the caption text rather than rigid judgment. Additionally, language ground truths, as lossy compressions of visual information, restrict the evaluation of more detailed captions. To improve applicability, we need to focus on the real ground truth---the image---and its fine-grained comprehension. 

These lead us to the CLIP score. However, the CLIP score has weaknesses. First, it only measures the consistency between the vision and language inputs but does not indicate whether the caption is undetailed or over-hallucinated. Moreover, the CLIP encoders focus on the main objects and overlook details, lacking a deep and fine-grained digest. \Cref{fig:clip_is_biased} shows an example where the CLIP image encoder omits the information in the background, leading to a lower CLIP score for the most detailed caption. In response, we propose a novel set of metrics: CLIP-Precision, CLIP-Recall, and CLIP-F1.

For an image-caption pair $(\bm{v},\bm{y})$, assume we have dense partitions of both vision and language. For the image, partitions are cropped sub-images of different objects or environments, including the full image. For the caption, partitions are sentences or phrases describing different objects or their characteristics, including the full caption. 
% Let $\mathcal{V}^{\text{p}}=\{\bm{v}^{\text{p}}_{i}\}_1^{n_v}$ represent the set of vision partitions and $\mathcal{Y}^{\text{p}}= \{ \bm{y}^{\text{p}}_{i}\}_1^{n_y}$ denote the set of language partitions, where $n_v$ and $n_y$ is the number of vision and language partitions, respectively.
Given these image and caption partitions, we utilize the CLIP image and text encoders to encode them, obtaining two sets of embeddings, both in various granularity levels. Let $\mathcal{V}^{\text{c}}=\{\bm{v}^{\text{c}}_{i}\}_1^{n_v}$ and $\mathcal{Y}^{\text{c}}= \{ \bm{y}^{\text{c}}_{i}\}_1^{n_y}$ denote the set of embeddings from image and caption partitions, respectively, where $n_v$ and $n_y$ represent the set sizes. We then compute the proportion that one embedding set can be linearly represented by the other as CLIP-Recall or CLIP-Precision and combine them to produce CLIP-F1. 

Consider an embedding $\bm{a} \in \mbR^{n_e}$ and a set $\mathcal{B} = \{\bm{b}_{i}\}_1^{n_b}$, where $n_e$ is the number of dimensions, $n_b$ is the size of $\mathcal{B}$. The proportion of top$k$-similar linear representation for $\bm{a}$ by $\mathcal{B}$ is defined as the $l_2$ norm ratio of the best linear combination of top$k$-similar embeddings in $\mathcal{B}$ to the embedding $\bm{a}$,
\begin{align*}
    &\text{PLR}_k(\bm{a}; \mathcal{B}) \coloneqq \frac{||B_k \omega^*||_2}{||\bm{a}||_2}, \\
    \st &\omega^* = \argmin_\omega ||B_k \omega - \bm{a}||_2,
\end{align*}
where $k$ is a parameter, $B_k \in \mbR^{n_e \times \min (k, n_b)}$ is a matrix whose columns are the embeddings $\bm{b}_{i} \in \mathcal{B}$ with the highest similarities to $\bm{a}$, and $||\cdot||_2$ is the $l_2$ norm. Notably, when we set $k=1$, or there is only one embedding in $\mathcal{B}$, i.e., $n_b=1$, the proportion of linear representation generalizes \textit{cosine similarity}. We discuss the effect of $k$ in \Cref{apdx:clip_metric_set_subsec}.

To obtain the recall, for each $\bm{v}^{\text{c}}_{i} \in \mathcal{V}^{\text{c}}$, we form a matrix $Y^{\text{c}}_k$ using the $\min (k, n_v)$ embeddings of caption partitions in $\mathcal{Y}^{\text{c}}$ with the highest similarities to $\bm{v}^{\text{c}}_{i}$ and compute the proportion of linear representation. We then compute the CLIP-Recall as the average PLR by $Y^{\text{c}}_k$ across $\mathcal{V}^{\text{c}}$:
\begin{equation*}
    \setlength\abovedisplayskip{3pt}
    \setlength\belowdisplayskip{3pt}
    \text{CLIP-Recall} = \frac{1}{n_v} \sum_{i=1}^{n_v} \text{PLR}_k(\bm{v}^{\text{c}}_{i}; \mathcal{Y}^{\text{c}}).
\end{equation*}
Similarly, we define CLIP-Precision as the average proportion of top$k$-similar linear representation by the image partitions across the caption partitions:
\begin{equation*}
    \setlength\abovedisplayskip{3pt}
    \setlength\belowdisplayskip{3pt}
    \text{CLIP-Precision} = \frac{1}{n_v} \sum_{i=1}^{n_v} \text{PLR}_k(\bm{y}^{\text{c}}_{i}; \mathcal{V}^{\text{c}}).
\end{equation*}

Intuitively, CLIP-Recall shows how much information in the image can be explained by the caption in various granularity levels, and CLIP-Precision measures the reverse. Thus, a higher CLIP-Recall score indicates a more detailed caption, while a lower CLIP-Precision score indicates more hallucinations. Finally, we combine CLIP-Recall and CLIP-Precision to obtain a general measure of how well the caption aligns with the image:
\begin{equation*}
    \setlength\abovedisplayskip{3pt}
    \setlength\belowdisplayskip{3pt}
    \text{CLIP-F1} \coloneqq \frac{2\text{CLIP-Recall} \cdot \text{CLIP-Precision}}{\text{CLIP-Recall}+\text{CLIP-Precision}}.
\end{equation*}

\begin{table*}[]
\centering
\label{tab:clip_metric_set}
\resizebox{\linewidth}{!}{
\begin{tabular}{cc|ccc|ccc|ccc}
\toprule
\multicolumn{2}{c|}{Model}                                            & \multicolumn{3}{c|}{MiniGPT-4}                                                                                  & \multicolumn{3}{c|}{LLaVA-1.5}                   & \multicolumn{3}{c}{mPLUG-Owl2}                   \\ \hline
\multicolumn{1}{c|}{Method}                                      & k  & CLIP-Recall                            & CLIP-Precision                & CLIP-F1                                & CLIP-Recall    & CLIP-Precision & CLIP-F1        & CLIP-Recall    & CLIP-Precision & CLIP-F1        \\ \hline
\multicolumn{1}{c|}{Greedy}                                      &    & 25.34                                  & 25.84                         & 25.53                                  & 25.47          & 28.31          & 26.77          & 25.21          & 27.62          & 26.32          \\
\multicolumn{1}{c|}{Beam}                                        &    & 24.97                                  & 26.22                         & 25.53                                  & 25.59          & 28.38          & 26.87          & 25.37          & 27.77          & 26.48          \\
\multicolumn{1}{c|}{VCD}                                         & 3  & 25.29                                  & 25.80                         & 25.49                                  & 25.50          & 28.37          & 26.82          & 25.22          & 27.60          & 26.32          \\
\multicolumn{1}{c|}{Opera}                                       &    & 24.70                                  & \textbf{26.30}                & 25.43                                  & 25.43          & \textbf{28.49} & 26.84          & 25.10          & 27.81          & 26.33          \\
\rowcolor[HTML]{EFEFEF} 
\multicolumn{1}{l|}{\cellcolor[HTML]{EFEFEF}\textbf{DBD (Ours)}} &    & \cellcolor[HTML]{EFEFEF}\textbf{26.16} & \cellcolor[HTML]{EFEFEF}25.69 & \cellcolor[HTML]{EFEFEF}\textbf{25.87} & \textbf{25.79} & 28.40          & \textbf{26.98} & \textbf{25.40} & \textbf{27.98} & \textbf{26.58} \\ \hline
\multicolumn{1}{c|}{Greedy}                                      &    & 26.20                                  & 26.26                         & 26.17                                  & 26.07          & 28.78          & 27.31          & 25.86          & 28.07          & 26.88          \\
\multicolumn{1}{c|}{Beam}                                        &    & 25.77                                  & 26.65                         & 26.14                                  & 26.18          & 28.82          & 27.39          & 26.02          & 28.22          & 27.03          \\
\multicolumn{1}{c|}{VCD}                                         & 5  & 26.14                                  & 26.22                         & 26.12                                  & 26.11          & 28.84          & 27.35          & 25.86          & 28.04          & 26.87          \\
\multicolumn{1}{c|}{Opera}                                       &    & 25.53                                  & \textbf{26.73}                & 26.06                                  & 26.01          & \textbf{28.95} & 27.36          & 25.83          & 28.35          & 26.98          \\
\rowcolor[HTML]{EFEFEF} 
\multicolumn{1}{l|}{\cellcolor[HTML]{EFEFEF}\textbf{DBD (Ours)}} &    & \cellcolor[HTML]{EFEFEF}\textbf{27.08} & \cellcolor[HTML]{EFEFEF}26.10 & \cellcolor[HTML]{EFEFEF}\textbf{26.52} & \textbf{26.48} & 28.87          & \textbf{27.57} & \textbf{26.07} & \textbf{28.43} & \textbf{27.15} \\ \hline
\multicolumn{1}{c|}{Greedy}                                      &    & 27.14                                  & 27.32                         & 27.15                                  & 26.40          & 29.94          & 28.00          & 26.29          & 29.24          & 27.63          \\
\multicolumn{1}{c|}{Beam}                                        &    & 26.60                                  & 27.75                         & 27.10                                  & 26.55          & 29.99          & 28.11          & \textbf{26.52} & 29.38          & 27.83          \\
\multicolumn{1}{c|}{VCD}                                         & 10 & 27.06                                  & 27.28                         & 27.09                                  & 26.42          & 30.01          & 28.04          & 26.29          & 29.21          & 27.63          \\
\multicolumn{1}{c|}{Opera}                                       &    & 26.42                                  & \textbf{27.81}                & 27.03                                  & 26.31          & 30.14          & 28.05          & 26.07          & 29.40          & 27.57          \\
\rowcolor[HTML]{EFEFEF} 
\multicolumn{1}{l|}{\cellcolor[HTML]{EFEFEF}\textbf{DBD (Ours)}} &    & \textbf{28.27}                         & 27.12                         & \textbf{27.61}                         & \textbf{26.66} & \textbf{30.15} & \textbf{28.22} & 26.27          & \textbf{29.69} & \textbf{27.87} \\ \bottomrule
\end{tabular}
}
\caption{Proposed CLIP metric sets evaluation results on the Visual Genome dataset of LVLMS with different methods. Bold indicates the best results of all methods. $k$ is the parameter for the CLIP metric set. 
For all $k$, higher CLIP-Recall indicates more details in captions; lower CLIP-Precision represents more hallucination.
}
\vspace{-4mm}
\end{table*}

\section{Evaluation}

\begin{table}[]
\centering
\label{tab:ablation_study}
\resizebox{0.95\linewidth}{!}{
\begin{tabular}{c|c|c|c|c}
\toprule
Model      & $\alpha_t$ & CLIP-Recall    & CLIP-Precision & CLIP-F1        \\ \hline
           & 2     & 27.06          & 26.08          & 26.50          \\
MiniGPT-4  & 3     & 27.00          & 25.96          & 26.42          \\
           & 4     & \textbf{27.08} & \textbf{26.10} & \textbf{26.52} \\ \hline
           & 2     & 26.45          & \textbf{28.87} & 27.55          \\
LLaVA-1.5  & 3     & 26.39          & 28.86          & 27.51          \\
           & 4     & 26.42          & 28.82          & 27.51          \\
           & 5     & \textbf{26.48} & \textbf{28.87} & \textbf{27.57} \\ \hline
           & 2     & 25.98          & 28.36          & 27.07          \\
mPLUG-Owl2 & 3     & 26.05          & 28.42          & 27.13          \\
           & 4     & \textbf{26.10} & 28.39          & 27.14          \\
           & 5     & 26.07          & \textbf{28.43} & \textbf{27.15}                          \\ \bottomrule
\end{tabular}
}
\caption{Proposed CLIP metric sets evaluation results on DBD with different $\alpha_t$. Higher $\alpha_t$ represents a stronger influence from the differential score on the decoding.}
\vspace{-3mm}
\end{table}

\subsection{Experiment Setting}

\noindent\textbf{LVLMs and dataset}. We evaluate the relationship between details and hallucinations on three state-of-the-art LVLMs, LLaVA-1.5~\cite{liu2023improved-llava15}, mPLUG-Owl2~\cite{ye2023mplugowl2}, and Minigpt-4-v2~\cite{chen2023minigptv2}. Our experiments are conducted on the Visual Genome dataset~\cite{krishna2017VG}, from which we randomly pick $500$ images from the validation set using the same random seed across different methods and LVLMs. 
% We report the average and standard deviation of all runs. 

\noindent\textbf{Baselines}. To validate our method, we compare it against classic decoding baselines such as greedy decoding and beam search, as well as state-of-the-art approaches to mitigate object hallucination in image captioning, including VCD~\citep{leng2023mitigating} and Opera~\citep{huang2024opera}. Details on baseline settings are provided in \Cref{apdx:experiment_subsec}.

\noindent\textbf{Metric implementation details}. We employ our proposed metrics—CLIP-Recall, CLIP-Precision, and CLIP-F1—for evaluation. We utilize the ViT-L/14 CLIP image and text encoder for these metrics, which produces the embeddings of $768$ dimensions. The Visual Genome dataset annotates dense regions and objects within each image with specific bounding boxes. We utilize these bounding boxes of regions and objects as the vision partitions of the images. Additionally, each image is treated as a special partition in its entirety.
% To supplement this, we randomly select and include additional $5$ regions or bounding boxes with random locations and random sizes for each image. 
For language partitions, we segment the captions into sentences based on punctuation marks like periods, question marks, and exclamation points and include the full caption as well.
% In experiments using our proposed method, DBD, we also test the case of different unit facts as the language partitions. 
We compute the CLIP metric set with different parameters $k$ and report all results.

\noindent\textbf{Method implementation details}. In the search phase, for LLaVA-1.5 and mPLUG-Owl2, we set the differential weight $\alpha_t$ as $10$ in the first $3$ decoding step, i.e., $t \le 3$, reducing it to 5 subsequently. For MiniGPT-4, we set $\alpha_t=4$ in all steps. We configure LLaVA-1.5 and mPLUG-Owl2 to use $5$ differentiated beams with a top$k$ setting of $6$, while MiniGPT-4 uses $7$ differentiated beams with a top$k$ of $7$. In the selection phase, we set a differential selection weight $\Bar{\alpha}$ as $5$ and the number of final unit facts $\Bar{n}$ as $10$. For more implementation details, please refer to the \Cref{apdx:impl_det_subsec}.

\subsection{Experiment Results}
We present the evaluation results of our proposed CLIP metric set for our method, Differentiated Beam Decoding (DBD), and various baselines across different LVLMs, as compiled in \Cref{tab:clip_metric_set}. The results consistently show that our DBD method surpasses all baseline methods. Compared to Greedy, Beam Search, and VCD, our method demonstrates superior CLIP-Recall and CLIP-Precision, indicating that DBD effectively enhances detail in LVLM-based image captioning without increasing the hallucination rate. While Opera Opera occasionally achieves better results on CLIP-Precision, this is paid for by its generally lower scores on CLIP-Recall and CLIP-F1.

Additionally, we report the ablation study on differential score weight $\alpha_t$ in \Cref{tab:ablation_study}. The results illustrate that while variations in $\alpha_t$ impact performance, DBD remains robust and not overly sensitive to this hyperparameter. Detailed caption comparison examples from the Visual Genome dataset are provided in \Cref{apdx:experiment_subsec}.

\section{Limitations}
While our method (DBD)  effectively mitigates the problem of extra interference from pre-generated texts, it does not address other potential sources of hallucinations. However, the unique parallel decoding structure of DBD offers the potential for integration with most existing hallucination-mitigating methods, a synergy that remains to be explored. Additionally, our proposed CLIP metric set relies on given image partitions. This requirement constrained our experiments to Visual Genome, which, despite its comprehensiveness, may not capture the full diversity of real-world scenarios.

\section{Conclusion}
In this paper, we address the debate on whether more detailed descriptions in LVLM-based image captioning necessarily lead to increased object hallucinations. We propose a novel decoding method, Differentiated Beam Decoding (DBD), accompanied by a new set of metrics: CLIP-Recall, -Precision, and -F1. Extensive experiments on the Visual Genome dataset validate that our method produces more detailed and accurate image captions while effectively maintaining a low hallucination level. Future research can extend our methods to further improve the robustness and reliability of LVLM-based applications across diverse scenarios.

\bibliography{reference}
\clearpage
\appendix

\section{Appendix}
\label{sec:appendix}

\subsection{Detail-Hallucination Debate} \label{apdx:det_hall_debate}
\noindent\textbf{Co-occurrence of Detail and Hallucination}. To fully support the co-occurrence side's arguments, we show that LVLMs tend to describe details in the later part of captions by a simple experiment. Concretely, we first generate captions of 500 images from the Visual Genome dataset using different LVLMs. Then, we match the object words in the captions with the object annotations in the dataset and compute the object size using the corresponding bounding box annotations. We report how the average object size changes depending on where it appears in the caption. As presented in the \Cref{fig:det_app_later}, the average size of objects consistently decreases as the sentence progresses, validating the statement.

One may find the co-occurrence of details and hallucinations similar to the natural phenomenon that "the more you say, the more mistakes you make." However, they are different. The former is saying that when describing details, it's more likely to produce hallucinations, and the latter is still tenable even if the hallucination frequency is common across all positions in the description.

\begin{figure*}[t]
    \centering
    \includegraphics[width=\linewidth]{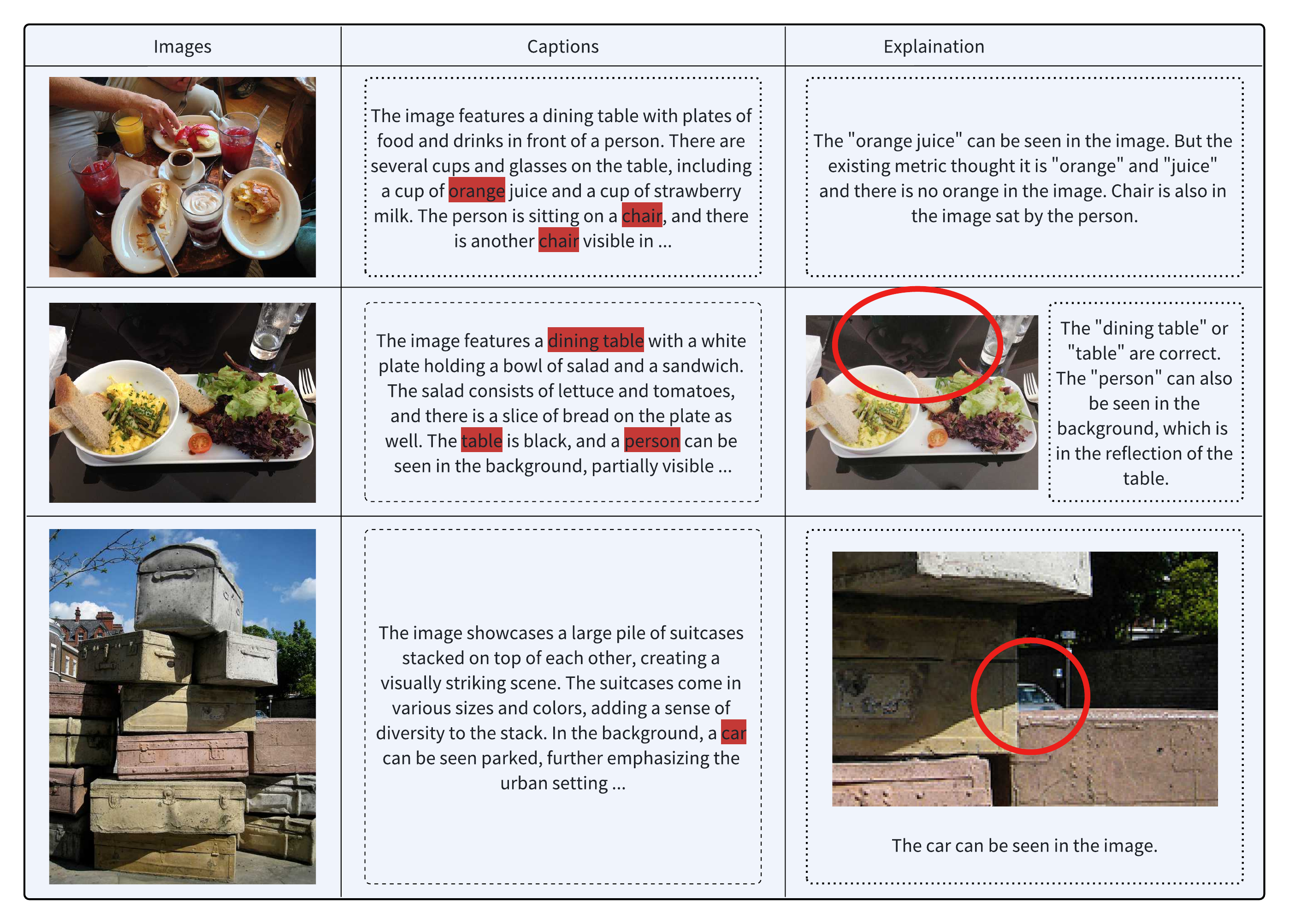}
    \caption{Examples illustrating misjudgments in image captioning. The left column contains images, the middle column shows automatically generated captions with highlighted sections in red that are considered incorrect by existing metrics (hallucination words). However, our model's captions are actually correct. The right column provides explanations and clarifications, demonstrating the accuracy of the highlighted sections and the flaws of the previous metric.}
    \label{fig:appendix_1}
    % \vspace{-6mm}
\end{figure*}

\begin{figure*}[t]
    \centering
    \includegraphics[width=\linewidth]{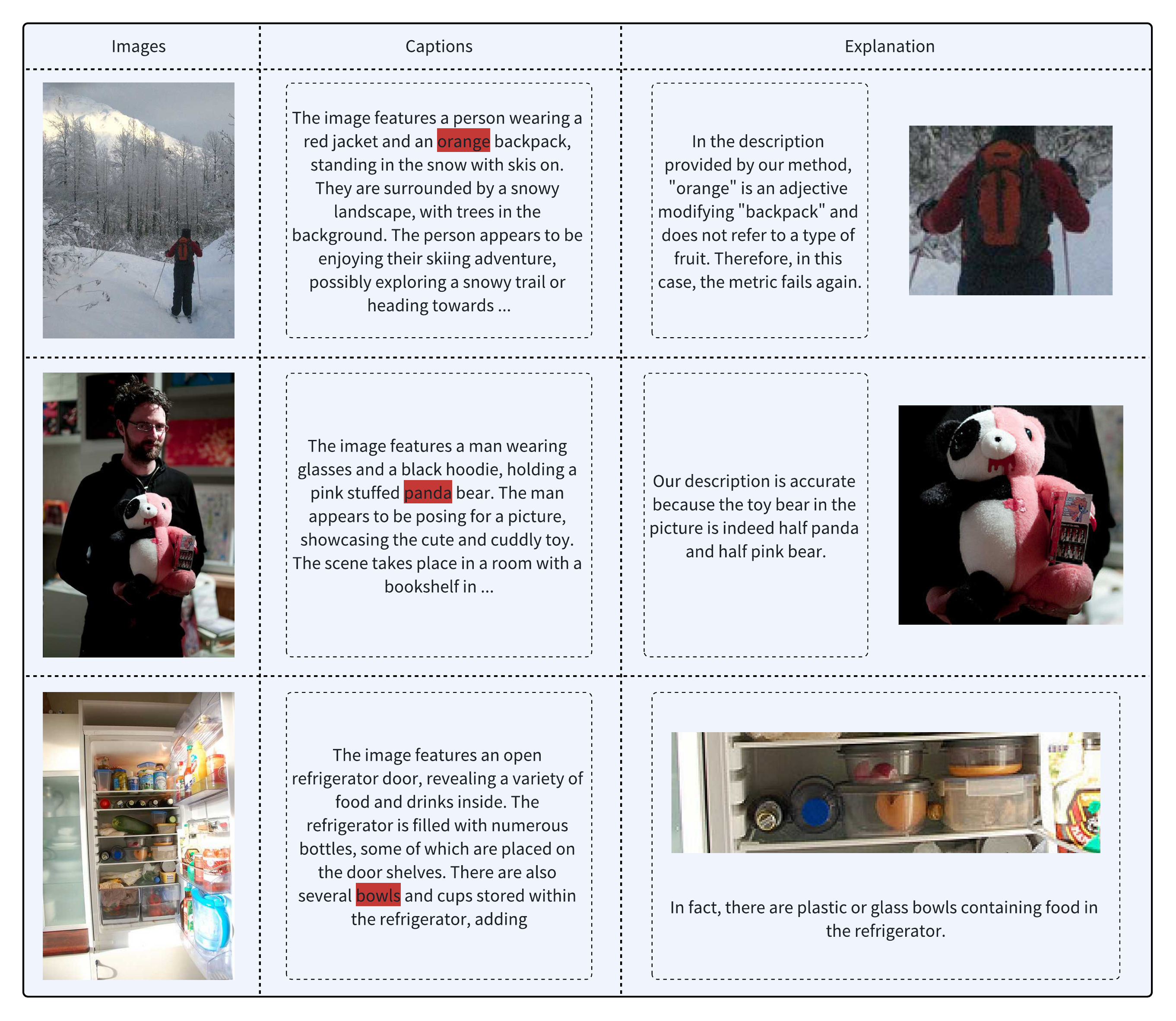}
    \caption{More examples illustrating misjudgments in image captioning.}
    \label{fig:appendix_2}
    % \vspace{-6mm}
\end{figure*}

\noindent\textbf{Inflated Hallucination Presence}. We provide more examples of failures in the current hallucination evaluation framework in \Cref{fig:appendix_1} and \Cref{fig:appendix_2}. Images are from the MSCOCO validation set. The captions are generated by LLaVA-1.5 using our proposed decoding method, DBD.

\subsection{Methodology: DBD}

\subsubsection{Preliminaries}
\noindent\textbf{LVLMs}. The architecture of the LVLM $\mathcal{M}$ incorporates a visual encoder, a cross-modality aligner, and a Large Language Model (LLM) core structure. Initially, the input undergoes processing by the visual encoder, followed by the cross-modality aligner to obtain a visual embedding in the textual space. Subsequently, the LLM core utilizes this visual-textual embedding, along with the text prompt and previously generated tokens, to predict the probability of the next token. 

\subsubsection{Discussion} \label{apdx:disc_beam_subsection}

\noindent\textbf{Beam Search}. Our approach is grounded in Beam Search (BS),  which aims to select an output sequence that maximizes the total log-likelihood, not just the highest probability next token at each step:
\begin{equation*} \label{eq:beam_search}
    \setlength\abovedisplayskip{3pt}
    \setlength\belowdisplayskip{3pt}
    \bm{y}_{\text{Beam}} = \argmax_{\bm{y} \in S^{\Bar{L}}} \sum_{t=1}^{\Bar{L}} \log p_{\mathcal{M}}(y_t \mid \bm{v},\bm{x},\bm{y}_{<t}),
\end{equation*}
where $\Bar{L}$ denotes the maximum output length and $\log p_{\mathcal{M}}(\bm{y}\mid \bm{v},\bm{x})$ represents the sentence-level log-likelihood. BS maintains $N_{\text{beam}}$ candidate sequences, each with a corresponding beam score, i.e., log-likelihood. In each step, BS calculates the top $K$ probable next tokens for each candidate sequence, resulting in $N~{\text{beam}}~K$ sub-candidates and their new log-likelihood scores. The top $N_{\text{beam}}$ sequences with the highest scores are then selected to continue to the next step.

\noindent\textbf{Relationship with Beam Search}.To eliminate the extra interference from pre-generated text in typical strategies, an intuitive approach is to process several separate output candidates simultaneously so that there is no precedence. While this sounds similar to Beam Search (BS), the candidates in conventional BS usually converge to similar full descriptions and focus on the same aspects of the image. Moreover, within each candidate, the generation of later texts is still influenced by earlier ones. Our method, however, encourages the exploration of distinct narrative directions in parallel and concludes when a single unit fact is completed.

\subsubsection{Implementation Details} \label{apdx:impl_det_subsec}
\noindent\textbf{Direct instruction for random aspects}. The inherent language priors of LVLMs often manifest a preference for describing certain aspects of an image over others, typically favoring a progression from central to peripheral elements or from general to specific details. This bias can impede the generation of differentiated facts, particularly when using a standard prompt such as \textit{"Please describe the image"}. To circumvent this issue, we leverage the LVLM's robust generalization capability by modifying our prompt during the differentiated beam decoding phase.

Specifically, we enhance the prompt to clarify the final goal of image captioning while also directly instructing the LVLM to focus on a randomly selected aspect of the image. This strategy encourages the model to diverge from its typical pattern of description, facilitating the generation of varied and independent factual content. For LLaVA-1.5 and mPLUG-Owl2, we use \textit{"Please generate a random fact of the image. You can describe the main object, the background, the environment, or any other detail. Please make sure the choice of the fact is random. Do not only focus on the people or the main object."} For MiniGPT-4, we add another \textit{"Give me the fact directly without other words"} at the end of the prompt to prevent MiniGPT-4 from generating too much nonsense. By directing the LVLM to select aspects randomly, we promote a more equitable consideration of all parts of the image, thereby supporting the production of a comprehensive and balanced description.

In the summarization phase, we prompt the \textit{"These are the image and the facts of the image. Please summarize them and generate a detailed description of the image based on the facts and the image"} for LLaVA-1.5 and mPLUG-Owl2. Similarly, we add a \textit{"Give me the fact directly without other words"} at the end.

\subsection{Evaluation}

\subsubsection{CLIP Metric Set} \label{apdx:clip_metric_set_subsec}
\noindent\textbf{Hyperparameter $k$}. The value of the parameter k represents how many of the most relevant embeddings can be used to fit an embedding from another modal. The design of the parameter $k$ is based on the observation that a description of an object or region may not be fully captured in just one sentence. A detailed caption may need to cover different aspects of the same area or object, leading to the use of multiple sentences. This requires allowing for a more accurate approximation of visual facts. Similarly, the evaluation of an overall description of a larger region may require the coordination of multiple dense image partitions. A higher value of $k$ would provide more flexibility in combining different partitions. However, setting $k$ too high may lead to interference from irrelevant partitions.

\subsubsection{Experiments} \label{apdx:experiment_subsec}
\noindent\textbf{Baseline settings}. We provide the hyperparameter you used in the experiments for baselines. For greedy search, there is no hyperparameter. For beam search, we set the number of beams as $5$ and the top$k$ as $5$. For VCD~\citep{leng2023mitigating}, we set $\alpha = 1$, $\beta=0.1$, and $t_{\text{noise}}=500$. For Opera~\citep{huang2024opera}, we set the number of beams as $5$, the top$k$ as $5$, the scale factor as $50$, the threshold as $15$, the number of attention candidates as $5$, and the penalty weight as $1$.
We prompt all baselines with “Please describe this image in
detail.”

\begin{figure*}[t]
    \centering
    \includegraphics[width=\linewidth]{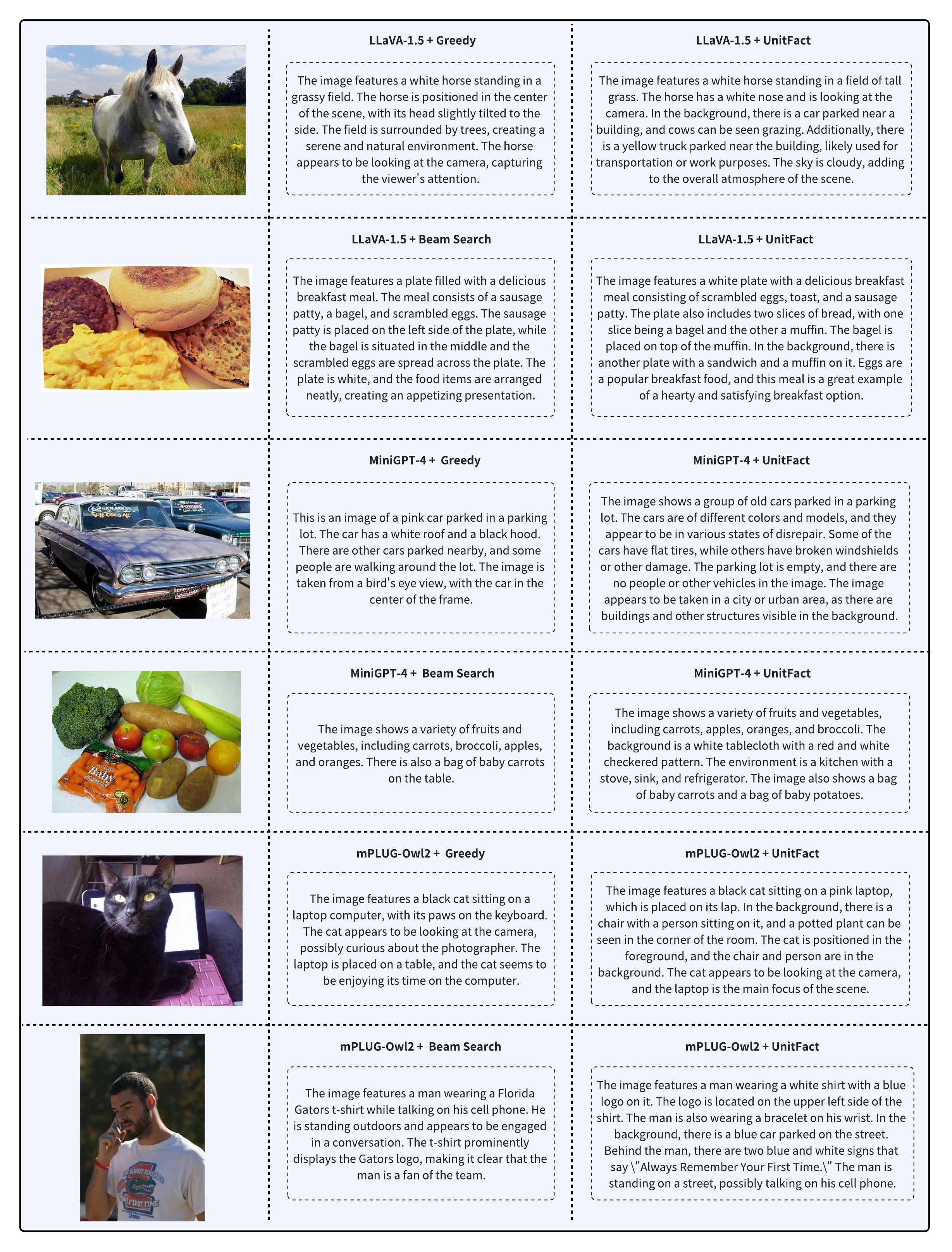}
    \caption{Comparison of image captioning results using different models and decoding methods: LLAVA-1.5 with Greedy and Beam Search, MiniGPT-4 with Greedy and UnitFact, and mPLUG-Owl2 with Greedy and UnitFact. The captions illustrate the differences in the models' abilities to accurately describe the content of the images, with UnitFact generally providing more detailed and contextually rich descriptions compared to Greedy and Beam Search strategies.}
    \label{fig:appendix_3}
    \vspace{-6mm}
\end{figure*}

\noindent\textbf{Caption examples}. We provide examples of examples of descriptions generated by our method on images from the Visual Genome dataset in \Cref{fig:appendix_3}.

\end{document}